\def\year{2021}\relax
\documentclass[letterpaper]{article} 
\usepackage{arxiv}  
\usepackage{times}  
\usepackage{helvet} 
\usepackage{courier}  
\usepackage[hyphens]{url}  
\usepackage{graphicx} 
\usepackage{natbib}  
\usepackage{caption} 

\usepackage{amsthm}
\newtheorem{theorem}{Theorem}
\newtheorem{lemma}{Lemma}

\usepackage{amssymb}
\usepackage{amsmath}
\usepackage{bm}
\usepackage{multirow}
\usepackage{subfigure}
\usepackage{color}

\title{High-Order Relation Construction and Mining for Graph Matching}
\author {
    Hui Xu,
    Liyao Xiang,
    Youmin Le,
    Xiaoying Gan,
    Yuting Jia,
    Luoyi Fu,
    Xinbing Wang\\
}
\affiliations{
    Shanghai Jiao Tong University \\
    \{xhui\_1,xiangliyao08,leyouming,ganxiaoying,hnxxjyt,yiluofu,xwang8\}@sjtu.edu.cn
}

\begin{document}

\maketitle

\begin{abstract}
Graph matching pairs corresponding nodes across two or more graphs. The problem is difficult as it is hard to capture the structural similarity across graphs, especially on large graphs. We propose to incorporate high-order information for matching large-scale graphs. Iterated line graphs are introduced for the first time to describe such high-order information, based on which we present a new graph matching method, called High-order Graph Matching Network (HGMN), to learn not only the local structural correspondence, but also the hyperedge relations across graphs. We theoretically prove that iterated line graphs are more expressive than graph convolution networks in terms of aligning nodes. By imposing practical constraints, HGMN is made scalable to large-scale graphs. Experimental results on a variety of settings have shown that, HGMN acquires more accurate matching results than the state-of-the-art, verifying our method effectively captures the structural similarity across different graphs.
\end{abstract}

\section{Introduction}

Graph matching refers to pairing corresponding nodes across two or more graphs by considering graph structural similarities and the optional attribute similarities. Since graphs are natural representations for many types of real-world data, the technique of graph matching lies at the core of many applications. Examples include but not limit to: 2D/3D shape matching for visual tracking in computer vision \cite{wang2019learning}, user accounts linkage across different online social networks \cite{liu2016aligning,fu2020anonymizing}, and entity alignment in cross-lingual knowledge graphs \cite{wu2019relation,xu2019cross}.

High-order information has been proven useful in graph matching theoretically and empirically \cite{zass2008probabilistic,duchenne2011tensor}. Different from the first-order (node) and second-order (edge), the high-order information is associated with specified node sets which compose connected components of a graph, typically referred to as {\em hyperedges}. Since corresponding nodes often share similar structures in their neighborhoods, high-order information are helpful in capturing similarity across graphs and identifying the corresponding nodes more precisely. Previous works have tried to incorporate high-order information for achieving better matching accuracies. Some works \cite{duchenne2011tensor,yan2015discrete} formulate graph matching as high-order affinity tensor based model and solve it with optimization techniques. The concept of hypergraphs are introduced in the learning based framework \cite{tan2014mapping} for graph matching. Recently, {\em Graph Convolution Networks} (GCNs) are exploited in \cite{wang2018cross,xu2019cross} to aggregate the high-order neighborhood information. It is proven that GCNs learn the hyperedges' representations approximately.

\begin{figure}[t]
\centering
\includegraphics[height=1.7cm,width=0.65\linewidth]{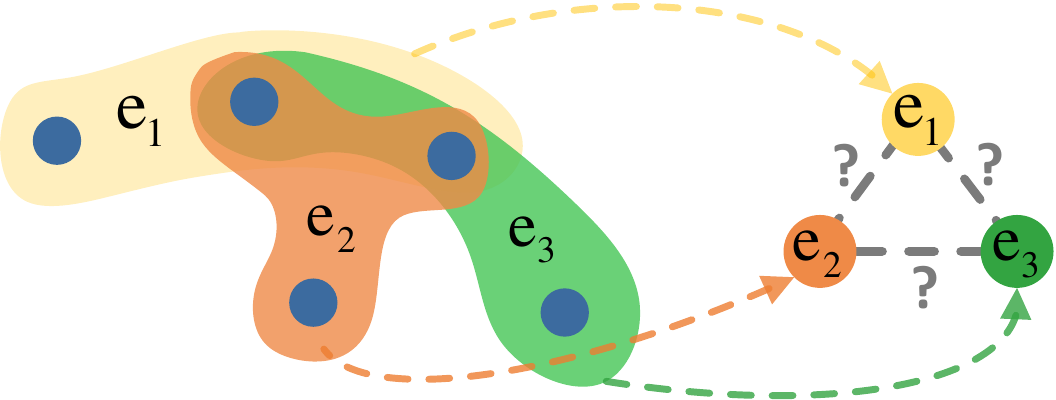}
\caption{An illustration of hypergraphs: an edge can join any number of nodes. Previous works do not explicitly consider the relation across hyperedges.}
\label{hyper}
\end{figure}

In spite of their promising performance, prior works have not proposed a principled way to describe the high-order information on graphs. Fig.\,\ref{hyper} gives a typical example of hypergraphs where $e_{1}, e_{2},e_{3}$ each represents a subset of nodes on the original graph. It is worth mentioning that previous methods with high-order information merely leverage the combination of features of the corresponding nodes within one hyperedge, but ignore the relative structural information across hyperedges. However, such information plays an essential role in the characterization of social networks and molecule graphs \cite{morris2019weisfeiler}.

We aim to introduce the high-order information to solve the large-scale graph matching problem by addressing the following challenges. {\em First}, since hypergraphs are not natural graph representations, there is no principled ways to extract hyperedges on graphs without additional node attributes, and to express the relation between hyperedges afterwards. {\em Second}, it is not clear how to build the relation between the similarity in hypergraphs and the similarity in the original graphs in a learning based framework. {\em Third}, from an engineering perspective, the consideration of the relation across hyperedges naturally introduces great complexity into the graph matching problem, and thus would be problematic to scale to large-size graphs.

To solve the first problem, we propose a learning-based graph matching method called {\em High-order Graph Matching Networks} (HGMN), taking advantage of the high-order structural information on graphs. We novelly use {\em line graphs}, in particular, {\em iterated line graphs} (ILG) \cite{harary1960some}, to describe the high-order structural information of a graph. Each node of the ILG can be viewed as a hyperedge of the original graph, and ILG requires no additional node attribute in its construction. Since the ILG is iteratively built from the original graph, hierarchical structural information can be described. By further taking into account the relative structural information of the ILG, we prove that GCNs cannot be more expressive than our method in aligning corresponding nodes across graphs theoretically and empirically. To resolve the second issue, we apply a general GNN to the ILG as to the original graphs for learning the embedding of each node (corresponding to the hyperedge), and establishing the high-order relations between graphs. Finally, by imposing constraints on the maximum degree of the graph, we are able to control the computational complexity of HGMN on large-scale graphs.

Highlights of contributions are as follows: 
\begin{itemize}
	\item To resolve graph matching, we introduce iterated line graphs to describe the high-order information, which is helpful in capturing the hierarchical structure of original graphs. We prove that GCNs are no more expressive than iterated line graphs in terms of aligning corresponding nodes. 
	\item A novel GNN-based high-order graph matching method called HGMN is proposed, which can utilize high-order structural similarity to get a more accurate matching result.
	\item Evaluated on a variety of real-world datasets in different settings, HGMN is shown to have superior performance than the state-of-the-art graph matching methods.
\end{itemize}

\section{Related Works}

\textbf{Graph matching} tries to find correspondences between two graphs and is conventionally formulated as Quadratic Assignment Problem (QAP) in \cite{loiola2007survey,cho2010reweighted,zhou2015factorized}. The problem is known as NP-complete and solved with optimization techniques. Works such as \cite{duchenne2011tensor,yan2015discrete, nguyen2015flexible} extend QAP to high-order tensor forms by encoding affinity between two hyperedges from graphs. However, the optimization-based approaches are not scalable to large-scale graphs.

There are many learning-based methods proposed to tackle graph matching, or network alignment, for large-scale inputs: IONE \cite{liu2016aligning} introduces a unified optimization framework to solve the network embedding and alignment tasks simultaneously. DeepLink \cite{zhou2018deeplink} samples the networks by random walk and introduces a dual learning method. CrossMNA \cite{chu2019cross} performs multi-graph alignment by using the cross-network information to refine the inter- and the intra- node embedding vectors respectively. Recently, many approaches leverage GCNs to capture the high-order information for graph matching, since GCNs is capable of aggregating $m$-hop neighborhood information of each node by stacking $m$ layers of GCN. GMNN \cite{xu2019cross} merges the node-level and the graph-level matching results by adopting GCN layers and cross-graph attention mechanisms. DGMC \cite{fey2020deep} presents a two-stage deep neural architecture for reaching a data-driven neighborhood consensus, and proposes optimization to fit to the large input domains. Although previous works have achieved some success, GCNs-based methods do not fully take advantage of the high-order structural information. We construct iterated line graphs as powerful expressions of high-order information in this paper.

Some works explicitly propose graph matching methods using \textbf{hypergraphs}. MAH \cite{tan2014mapping} is inspired by the intuition that nodes within a hyperedge should have higher similarity than the nodes belonging to different hyperedges. MGCN \cite{chen2020multi} is an enhanced version of MAH as it considers multi-level graph convolutions on both local network structures and hypergraphs in a unified way. In fact, MGCN focuses on the anchor link prediction problem, which is related to but different from the graph matching problem. Although both MAH and MGCN utilize hypergraphs, they are unable to describe graph structures of different granularities while our methods can establish hyperedges in a hierarchical way with enriched information.

\begin{figure}[t]
    \centering
    \includegraphics[height=2cm, width=0.8\linewidth]{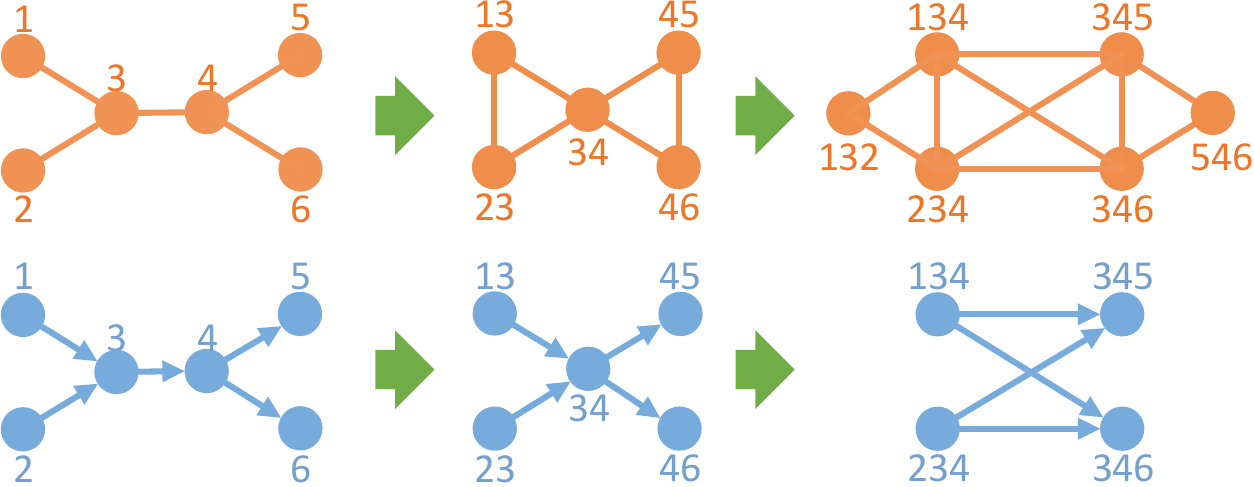}
    \caption{Construction of iterated line graphs for directed~(lower) and undirected~(upper) graphs.}
    \label{fig:ilg}
\end{figure}

\section{Preliminaries}

\subsubsection{Hypergraphs and Hyperedges.} Formally, a hypergraph $\mathcal{G}^h=(\mathcal{V}, \mathcal{E}^h)$ consists of a set of nodes $\mathcal{V}$, and a set of non-empty subsets of $\mathcal{V}$, namely hyperedges $\mathcal{E}^h$. For each hyperedge $e \in \mathcal{E}^h$, we have $e=\{v_1,...,v_k\}$, $v_i\in \mathcal{V}$, $2 \leq k \leq |\mathcal{V}|$.

\subsubsection{Line Graphs.} Suppose $\mathcal{G}=(\mathcal{V}, \mathcal{E})$ is a finite undirected graph, we denote $L(\mathcal{G}) = (\mathcal{V}_L, \mathcal{E}_L)$ as the line graph of $\mathcal{G}$, {\em i.e.,} the vertices of $L(\mathcal{G})$ are the edges of $\mathcal{G}$ and two vertices of $L(\mathcal{G})$ are adjacent if their corresponding edges in $\mathcal{G}$ have a node in common. The line graph of a directed graph $\mathcal{G}$ only has one difference compared to the undirected version such that, an edge in $L(\mathcal{G})$ is built only when the two corresponding directed edges $e_1$ and $e_2$ satisfy that the head of $e_1$ is the tail of $e_2$. Example constructions are shown in Fig.~\ref{fig:ilg}.

Incidence matrix $\bm{H} \in \mathbb{R}^{|\mathcal{V}| \times |\mathcal{V}_L|}$ builds the correspondence between $L(\mathcal{G})$ and $\mathcal{G}$ with each entry $\bm{H}(v, v_L)$ determined by:
\begin{equation}
\small
\bm{H}(v, v_L)=\left\{
\begin{aligned}
1&, \ \ \text{if} \ v ~\text{belongs to}\ e\ \text{(make}~ v_L) \\
0&, \ \ \text{otherwise.} \\
\end{aligned}
\right.
\end{equation}
We can construct a line graph out of a line graph in an iterated way. For simplicity, we express $m$-iterated line graph ($m$-ILG) $L(...(L(\mathcal{G})))$ as $L^m(\mathcal{G})$, and the incidence matrix is $\bm{H}^{(m)}=\bm{H}^{(0,1)}...\bm{H}^{(m-1,m)}$, where $\bm{H}^{(p,q)}$ links a $p$-ILG to a $q$-ILG.

\subsubsection{Graph Convolutional Networks.} Given a graph $\mathcal{G}$, the $(t+1)$-th layer of GCN aggregate the neighborhood information of each node via
\begin{equation} \label{eq:gcn}
\small
\bm{X}^{(t+1)}=\sigma \left(\bm{\tilde{A}}\bm{X}^{(t)}\bm{W}^{(t)}\right),
\end{equation}
where $\bm{X}$ is the node feature matrix, $\bm{\tilde{A}}=\bm{\tilde{D}}^{-\frac{1}{2}}(\bm{A}+\bm{I})\bm{\tilde{D}}^{-\frac{1}{2}}$ is the normalized adjacency matrix of $\mathcal{G}$ with self-connections, $\tilde{D}_{ii}=\sum_j(A_{ij}+1)$ is a diagnal matrix, and $\sigma(\cdot)$ is an activation function.

\section{Problem Definition}
Let $\mathcal{G}_s=(\mathcal{V}_s, \mathcal{E}_s, \bm{X_s}, \bm{E_s})$, $\mathcal{G}_t=(\mathcal{V}_t, \mathcal{E}_t, \bm{X_t}, \bm{E_t})$ be the source and target graph respectively, which consist of a finite set of nodes $\mathcal{V}=\{v_1,...,v_{|\mathcal{V}|}\}$, a finite set of edges $\mathcal{E}=\{e_{ij}\}_{i,j=1}^{|\mathcal{V}|}$, an optional node feature matrix $\bm{X} \in \mathbb{R}^{|\mathcal{V}| \times \cdot}$ and an optional edge feature matrix $\bm{E} \in \mathbb{R}^{|\mathcal{E}| \times \cdot}$. W.l.o.g., we assume that $|\mathcal{V}_s| \leq |\mathcal{V}_t|$.

The problem of graph matching is as follows. Given $\mathcal{G}_s$ and $\mathcal{G}_t$ with adjacency matrices $\bm{A}_s$ and $\bm{A}_t$ respectively, we aim to find a mapping matrix $\bm{S} \in \{0, 1\}^{|\mathcal{V}_s| \times |\mathcal{V}_t|}$ which follows one-to-one mapping constraints $\sum_{j \in \mathcal{V}_t} S_{i,j} \leq 1, \forall i \in \mathcal{V}_s$ and $\sum_{i \in \mathcal{V}_s} S_{i,j} \leq 1, \forall j \in \mathcal{V}_t$. We infer an injective mapping function $\pi : \mathcal{V}_s \to \mathcal{V}_t$ which maps each node in $\mathcal{G}_s$ to a node in $\mathcal{G}_t$. Conventionally, graph matching is expressed as an edge-preserving problem:
\begin{equation} \label{eqn:gm}
\bm{S} = \arg\min\left(||\bm{A}_s - \bm{S}^T\bm{A}_t\bm{S}||_F^2\right),
\end{equation}
subject to the one-to-one mapping constraints mentioned above. However, this formulation only leverages the local structure (edges) to find a mapping. In our work, we extend the local structural information to the high-order structural information by introducing iterated line graphs. One can interpret our problem as solving Eq.~\ref{eqn:gm} with the adjacency matrices replaced with expressions concerning iterated line graphs for better alignment on the original graphs.

\begin{figure*}[t]
	\centering
	\includegraphics[height=4.5cm, width=0.8\linewidth]{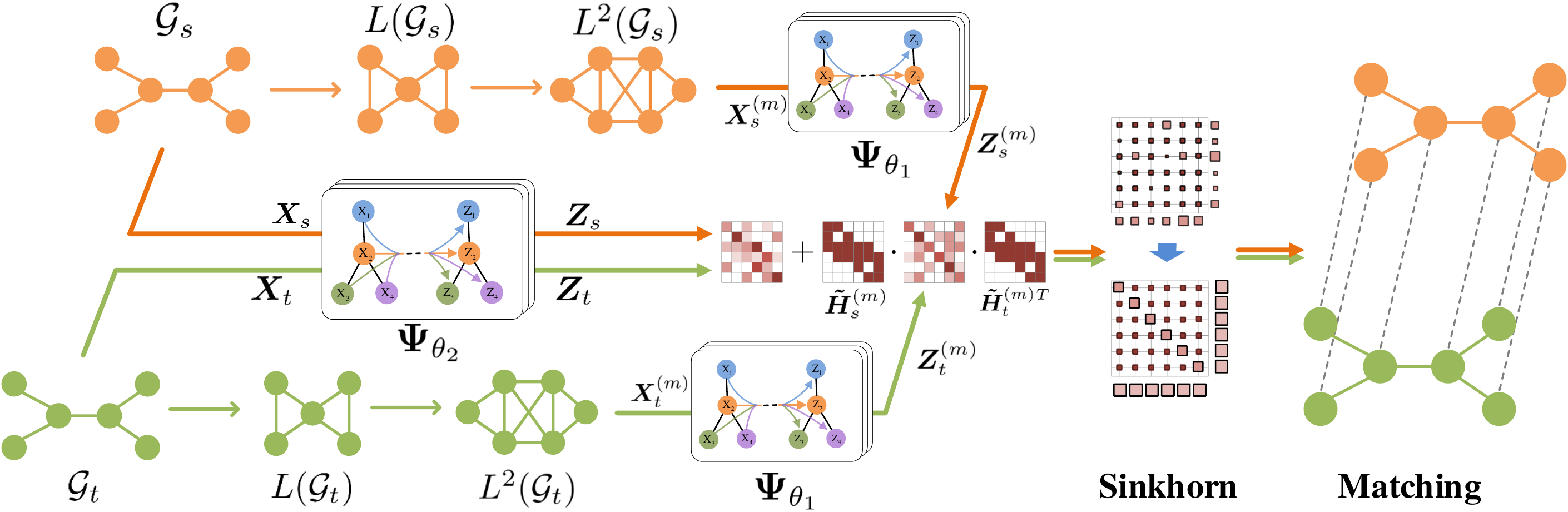}
	\caption{Overview of our high-order graph matching networks with iterated line graphs. Features of iterated line graphs and original graphs are learned by GNNs. Inner product is used to calculate the distance between node pairs and the weighted sum of the local and high-order corresponding matrices. Sinkhorn operations produce the one-to-one matching results.}
	\label{fig:mf}
\end{figure*}

\section{The Proposed Approach}
\subsection{Iterated Line Graph Construction}
Given the source graph $\mathcal{G}_s$ and the target graph $\mathcal{G}_t$, we construct the $m$-ILG $L^{m}(\mathcal{G})=(\mathcal{V}^{(m)}, \mathcal{E}^{(m)},\bm{A}^{(m)}, \bm{X}^{(m)})$ by the definition of ILGs. $\bm{A}^{(m)}$ and $\bm{X}^{(m)}$ are the adjacency matrix and the feature matrix of $L^{m}(\mathcal{G})$ respectively and can be constructed by setting $k$ to $\{1, \ldots,m \}$ iteratively in the following equations:
\begin{equation}
\small
\begin{aligned}
\bm{x}_i^{(k)} &= \textstyle{\bigoplus}_{j}(\bm{H}^{(k-1,k)}{}^T)_{i,j}\bm{x}_j^{(k-1)}, \\
\bm{A}^{(k)} &= \bm{H}^{(k-1,k)^T}\bm{H}^{(k-1,k)}-2\bm{I},
\end{aligned}
\label{concat}
\end{equation}
where $\oplus$ is the operation concatenating the non-zero vectors, and $\bm{X}^{(0)}=\bm{X}$ of the original graph.

Compared to the conventional hypergraph-based methods, we impose no additional information, {\em e.g.}, community or cluster information, to construct ILGs. Hence our methods are more general in expressing hypergraph structure without additional attributes of the graph. More importantly, the high-order structural information, such as the relation across different hyperedges, is explicitly expressed with ILGs, which is not captured in the previous definition of hypergraphs.

\subsection{High-order Graph Matching}
Given the $m$-ILGs of the source and the target graph, we implement Graph Neural Networks (GNNs) on $L^m(\mathcal{G}_s)$ and $L^m(\mathcal{G}_t)$ to learn the similarity between the two, which we referred to as $m$-order structural similarity. With the incidence matrices, we are able to project the $m$-order similarity to the similarity across the original source and target graph, expressed by $\bm{S}$:
\begin{equation}
\small
\begin{aligned}
\bm{S}=sinkhorn(\bm{\tilde{H}}_s^{(m)}\bm{Z}^{(m)}_s{\bm{Z}^{(m)}_t}{}^T{{{\bm{\tilde{H}}}_t}^{(m)}}{}^T&),
\end{aligned}
\label{m-order S}
\end{equation}
where $\bm{\tilde{H}}^{(m)} = \bm{D}_{\bm{H}^{(m)}}^{-1}\bm{H}^{(m)}$, $\bm{D}_{\bm{H}^{(m)}} \in \mathbb{R}^{|\mathcal{V}|\times|\mathcal{V}|}$ is a diagnoal matrix and $\bm{D}_{\bm{H}^{(m)}}(i,i)=\sum_jH^{(m)}_{i,j}$. $\bm{Z}^{(m)}_s=\mathbf{\Psi}_{\theta_1}(\bm{X}_s^{(m)}, \bm{A}_s^{(m)}, \bm{E}_s^{(m)})$ and $\bm{Z}^{(m)}_t=\mathbf{\Psi}_{\theta_1}(\bm{X}_t^{(m)}, \bm{A}_t^{(m)}, \bm{E}_t^{(m)})$ are ILG features computed from the shared GNN $\mathbf{\Psi}_{\theta_1}(\cdot)$. Sinkhorn normalization is applied to obtain rectangular doubly-stochastic correspondence matrices that fulfill the one-to-one mapping constraints $\sum_{j\in\mathcal{V}_t}S_{i,j}=1, \forall{i\in\mathcal{V}_s}$ and $\sum_{i\in\mathcal{V}_s}S_{i,j} \leq 1, \forall{j\in\mathcal{V}_t}$ \cite{sinkhorn1967concerning}. $\bm{\tilde{H}}^{(m)}\bm{Z}^{(m)}$ can be considered as features on the $m$-ILG projected to the original graph to match nodes in the original graphs.

In practice, we take the local structural information on the original graph into account, and calculate the similarity as a combination of the high-order and the local similarity:
\begin{equation}
\small
\begin{aligned}
\bm{S}=sinkhorn\Big(\alpha&(\underbrace{\bm{\tilde{H}}_s^{(m)}\bm{Z}^{(m)}_s{\bm{Z}^{(m)}_t}^T{\bm{\tilde{H}}_t^{(m)}}{}^T}_{high\textnormal{-}order\ similarity}) \\
&\ \ \ \ \ \ +(1-\alpha)\underbrace{(\bm{Z}_s{\bm{Z}_t}^T)}_{local\ similarity}\Big),
\end{aligned}
\label{HGMN}
\end{equation}
where $\bm{Z}_s=\mathbf{\Psi}_{\theta_2}(\bm{X}_s, \bm{A}_s, \bm{E}_s)$, and $\bm{Z}_t=\mathbf{\Psi}_{\theta_2}(\bm{X}_t, \bm{A}_t, \bm{E}_t)$. We train $\mathbf{\Psi}_{\theta_1}$ and $\mathbf{\Psi}_{\theta_2}$ by minimizing the cross entropy loss:
\begin{equation}\label{eq:loss}
\small
\mathcal{L}=-\sum_{i\in \mathcal{V}_s, j\in \mathcal{V}_t} S^{gt}_{i, j} \mathrm{log}(S_{i,j})
\end{equation}
where $S_{i,j}$ are the similarity to be learned and $S^{gt}_{i, j} = 1$ if and only if node $i\in \mathcal{G}_s$ has ground truth correspondence $j\in\mathcal{G}_t$.

We implement the siamese version of the GNN model, which has shown great advantage in capturing the similarity between different graphs \cite{li2019graph,fey2020deep}. Concretely, $\mathcal{G}_s$ and $\mathcal{G}_t$ share $\bm{\theta}_2$ of $\mathbf{\Psi}_{\theta_2}$ for learning local similarity, $L^{m}(\mathcal{G}_s)$ and $L^{m}(\mathcal{G}_t)$ share $\bm{\theta}_1$ of $\mathbf{\Psi}_{\theta_1}$ for learning high-order similarity. We implement $\mathbf{\Psi}_{\theta_1}, \mathbf{\Psi}_{\theta_2}$ as the standard GNNs, which generally aggregate the structural information and update the node features $\mathbf{x}_v^{(t-1)}$ in layer $t$ via
\begin{equation}
\small
\begin{aligned}
\mathbf{x}_{\mathcal{N}(v)}^{(t)} &= \mathrm{AGGREGATE}^{(t)}({\mathbf{x}_u^{(t-1)}, \forall u\in\mathcal{N}(v)}), \\
\mathbf{x}_v^{(t)} &= \mathrm{UPDATE}^{(t)}(\mathbf{x}_v^{(t-1)}, \mathbf{x}_{\mathcal{N}(v)}^{(t)}),
\end{aligned}
\end{equation}
where $\mathcal{N}(v)$ is the neighbor set of node $v$ on the original or any order of the ILGs. Fig.~\ref{fig:mf} shows our HGMN frameworks based on ILGs.

We will show how the ILG is related to the GCNs model with the Thm.~\ref{thm1}, and further display the unique advantage of our method in Thm.~\ref{thm2}.
\begin{theorem} \label{thm1}
The feature on the line graph is equivalently expressive as that of the one-layer GCN.
\end{theorem}

The proof of Thm.~\ref{thm1} can be found in the supplementary materials. We further found that ILGs are capable of describing high-order structural information which are missed by GCNs. Before introducing the Thm.~\ref{thm2}, we show the following lemma holds true:
\begin{lemma} \label{lem}
Let $\bm{A}$ be the adjacency matrix of graph $\mathcal{G}$, $\bm{A}^m$ is $\bm{A}$ raised to the $m$-th power, and $\bm{H}^{(m)}$ be the incidence matrix of $m$-ILG $L^{m}(\mathcal{G})$. We have $\mathbf{1}_{[(\bm{A}+\bm{I})^m > 0]} = \mathbf{1}_{[\bm{H}^{(m)}\bm{H}^{(m)}{}^{T} > 0]}$ where $\mathbf{1}_{[\bm{X} > 0]}$ represents that the element $x$ is set to $1$ where $x > 0$ in $\bm{X}$.
\end{lemma}

With Lemma~\ref{lem}, we can prove the following theorem:
\begin{theorem} \label{thm2}
ILGs are strictly more expressive than GCNs in expressing high-order structural information.
\end{theorem}

We have proven that $m$-layer GCNs without non-linear layers only capture the nodes within $m$-hop neighborhood of node $i\in\mathcal{G}$, without considering the relative positions of the nodes. While in $m$-ILG, the relations between different hyperedges are explicitly expressed and learned. Hence we think the ILG captures richer high-order structural information than GCNs.

\subsection{Hierarchical Variants}
Since Eq.~\ref{HGMN} only considers the $m$-th order similarity between $L^m(\mathcal{G}_s)$ and $L^m(\mathcal{G}_s)$, as well as the local similarity between $\mathcal{G}_s$ and $\mathcal{G}_t$, some crucial structural similarity information between $L^k(\mathcal{G}_s)$ and $L^k(\mathcal{G}_t) (1 \leq k < m)$, may be missed. Hence we propose a hierarchical variant of HGMN by training similarity $\bm{S}$ on the $k$-ILG from $k=0$ to $k=m$ independently and iteratively. After the training on the $(k-1)$-ILG is done, we initialize the feature of the $k$-ILG (source and target) as
\begin{equation}
\small
\bm{x}^{(k)}_i = \textstyle{\bigoplus}_{j}((\bm{H}^{(k-1,k)})^T)_{i,j}(\bm{Z}^{(k-1)})_j,
\label{variant}
\end{equation}
where $\bm{Z}^{(k-1)}$ is learned on the $(k-1)$-ILG. In the hierarchical variant, $k$-th order features are built iteratively. We do not train features of order $0$ to $m$ in an end-to-end fashion due to the high complexity issue.

\subsection{Complexity Analysis}
The time complexity of HGMN is determined by the complexity of GNN used. The sapce complexity is decided by sizes of ILGs. Assuming $d_{max}^{(m)}$ is the max degree of $m$-ILG $L^m(\mathcal{G})$, the upper bound of the size of $L^m(\mathcal{G})$ can be expressed as $|\mathcal{V}|\Pi_{l=0}^{m}(d_{max}^{(l)}/{2})$, given $\mathcal{G}$ with $|\mathcal{V}|$ nodes. To see it, we first obtain the upper bound of the number of edges of $\mathcal{G}$ as $|\mathcal{V}|(d_{max}^{(0)}/{2})$, which is also the upper bound of $|\mathcal{V}^{(1)}|$ in $L(\mathcal{G})$. The upper bound of $|\mathcal{V}^{(m)}|$ can be obtained iteratively. 

To calculate the corresponding matrix $\bm{S}$ in Eq.~\ref{HGMN}, it is required to store $\bm{\tilde{H}}^{(m)} \in \mathbb{R}^{|\mathcal{V}| \times |\mathcal{V}^{(m)}|}$, $\bm{Z}^{(m)} \in \mathbb{R}^{|\mathcal{V}^{(m)}| \times k}$, $\bm{\tilde{H}}^{(m)}\bm{Z}^{(m)} \in \mathbb{R}^{|\mathcal{V}| \times k}$, for both the source and target graph, where $k$ is the feature dimention of node in $L^m(\mathcal{G})$. Hence the space complexity of Eq.~\ref{HGMN} is $O(\sum_{i\in \{s, t\} }(|\mathcal{V}_i|^2\Pi_{l=0}^{m}(\frac{d_{max,i}^{(l)}}{2})+k|\mathcal{V}_i|\Pi_{l=0}^{m}(\frac{d_{max,i}^{(l)}}{2})+k|\mathcal{V}_i|)+|\mathcal{V}_s||\mathcal{V}_t|).$
Since the $k$-th order hierarchical variant is learned over the $(k-1)$-ILG, the space complexity depends on the highest order $m$. Such complexity is much smaller than that of high-order affinity tensor based method \cite{duchenne2011tensor}, which is $O((|\mathcal{V}_s||\mathcal{V}_t|)^m)$, on large graphs.

\subsection{Scaling To Large Graphs}
Although the complexity of HGMN is low compared to other schemes, it is still unacceptable for large-scale graphs. Hence we apply several optimization techniques to make HGMN more scalable in practice.

\textbf{Edge deletion.} According to the above complexity analysis, $d_{max}$ is an important factor to the space complexity. Hence we can control the complexity with a preset hyperparameter $d^{(k)}$ to constrain the size of the $k$-ILG. Specifically, to construct $L^{k+1}(\mathcal{G})$, we randomly select $\min\{|\mathcal{N}(v^{(k)})|, d^{(k)}\}$ neighbors of $v^{(k)}$ in $L^k(\mathcal{G})$ to keep the edges of $v^{(k)}$ connected to these neighbors and delete others. Note that we do not delete anything related to the nodes with ground truth as the information is crucial in training.

\textbf{Sparse correspondences.} Following the work of \cite{fey2020deep}, we also sparsify correspondence matrix $\bm{S}$ by filtering out the low rank correspondences. Concretely, we compute $\rm{Top}_k$ correspondences of each row $\bm{S}_{i, :}$ and store its sparse version including the ground truth entries $S_{i, \pi(i)}$. Although it still requires $O(|\mathcal{V}_s||\mathcal{V}_t|)$ to store the dense version of $\bm{S}$, the space consumption in the backpropagation stage is reduced by a large margin.

\begin{table}[tbp]
\centering
\caption{Real-world dataset used in our experiments, where $d_{max}$ is the max degree of each network.}
\resizebox{0.9\linewidth}{21mm}{
\begin{tabular}{cccccc}
\hline
\multicolumn{2}{c|}{Dataset} & \multicolumn{1}{c}{$|\mathcal{V}|$} & \multicolumn{1}{c}{$|\mathcal{E}|$} & \multicolumn{1}{c}{$d_{max}$}& \multicolumn{1}{c}{Seeds} \\
\hline
\multicolumn{1}{c|}{Twitter} & \multicolumn{1}{c|}{$\mathcal{G}_s$} & 5220 & 164919 & 1725 & \multirow{2}{*}{1609}\\
\multicolumn{1}{c|}{Foursquare} & \multicolumn{1}{c|}{$\mathcal{G}_t$} & 5315 & 76972& 552 & \\
\hline
\multicolumn{1}{c|}{AI} & \multicolumn{1}{c|}{$\mathcal{G}_s$}  & 12029 & 67760 & 116 &\multirow{2}{*}{1136}\\
\multicolumn{1}{c|}{DM} & \multicolumn{1}{c|}{$\mathcal{G}_t$}  & 8916 & 55112 & 145 &\\
\hline
\multicolumn{1}{c|}{AI$_{13,14}$} & \multicolumn{1}{c|}{$\mathcal{G}_s$}  & 7226 & 25081 & 73 & \multirow{2}{*}{2861}\\
\multicolumn{1}{c|}{AI$_{15,16}$} & \multicolumn{1}{c|}{$\mathcal{G}_t$}  & 10241 & 43534 & 73 &\\
\hline
\multicolumn{1}{c|}{$\mathrm{DBP}_{\mathrm{FR}}$} & \multicolumn{1}{c|}{$\mathcal{G}_s$}  & 19661 & 105998 & 145 &\multirow{2}{*}{15000}\\
\multicolumn{1}{c|}{$\mathrm{DBP}_{\mathrm{EN}}$} & \multicolumn{1}{c|}{$\mathcal{G}_t$}  & 19993 & 115722 & 142 &\\
\hline
\multicolumn{1}{c|}{$\mathrm{DBP}_{\mathrm{JA}}$} & \multicolumn{1}{c|}{$\mathcal{G}_s$}  & 19814 & 77214 & 76 &\multirow{2}{*}{15000}\\
\multicolumn{1}{c|}{$\mathrm{DBP}_{\mathrm{EN}}$} & \multicolumn{1}{c|}{$\mathcal{G}_t$}  & 19780 & 93484 & 135 &\\
\hline
\multicolumn{1}{c|}{$\mathrm{DBP}_{\mathrm{ZH}}$} & \multicolumn{1}{c|}{$\mathcal{G}_s$}  & 19388 & 70414 & 73 &\multirow{2}{*}{15000}\\
\multicolumn{1}{c|}{$\mathrm{DBP}_{\mathrm{EN}}$} & \multicolumn{1}{c|}{$\mathcal{G}_t$}  & 19572 & 95142 & 90 &\\
\hline
\end{tabular}
}
\label{tab1}
\end{table}

\begin{figure*}[t]
\setlength{\abovecaptionskip}{8pt}
\centering
\subfigure[Structural noise $p_d$]{
\centering
\includegraphics[width=0.23\linewidth]{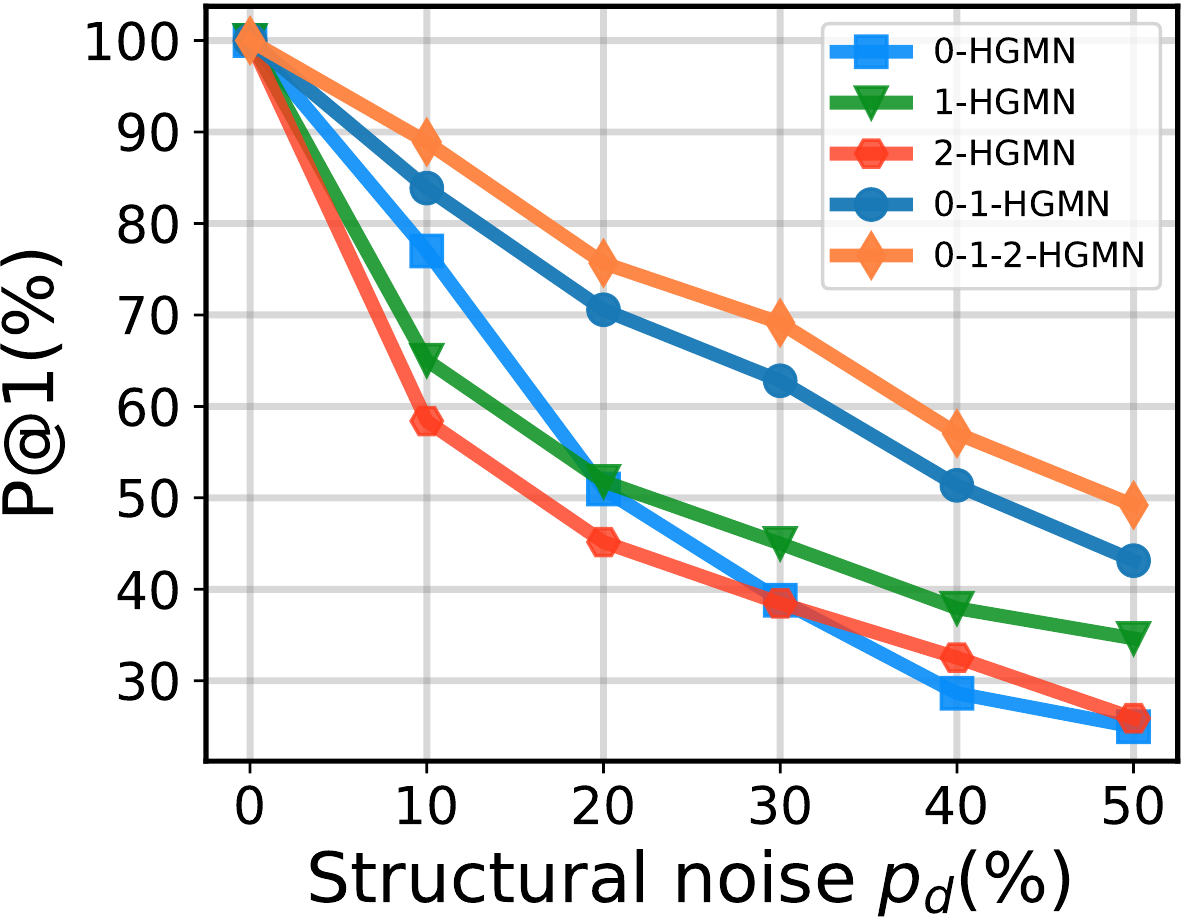}
}
\vspace{-0.2cm}
\subfigure[Num of layers $T$]{
\centering
\includegraphics[width=0.23\linewidth]{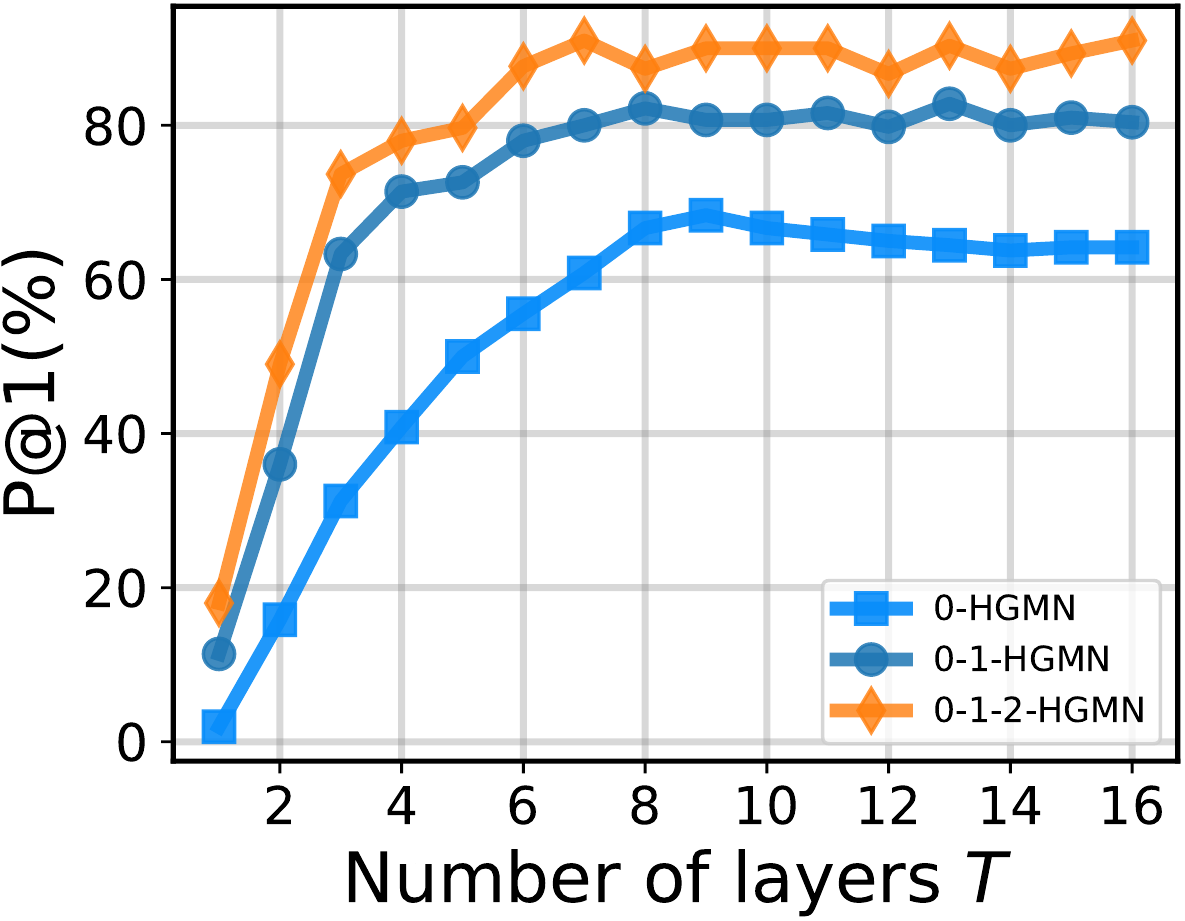}
}
\vspace{-0.2cm}
\subfigure[Hyperparameters $\alpha$]{
\centering
\includegraphics[width=0.23\linewidth]{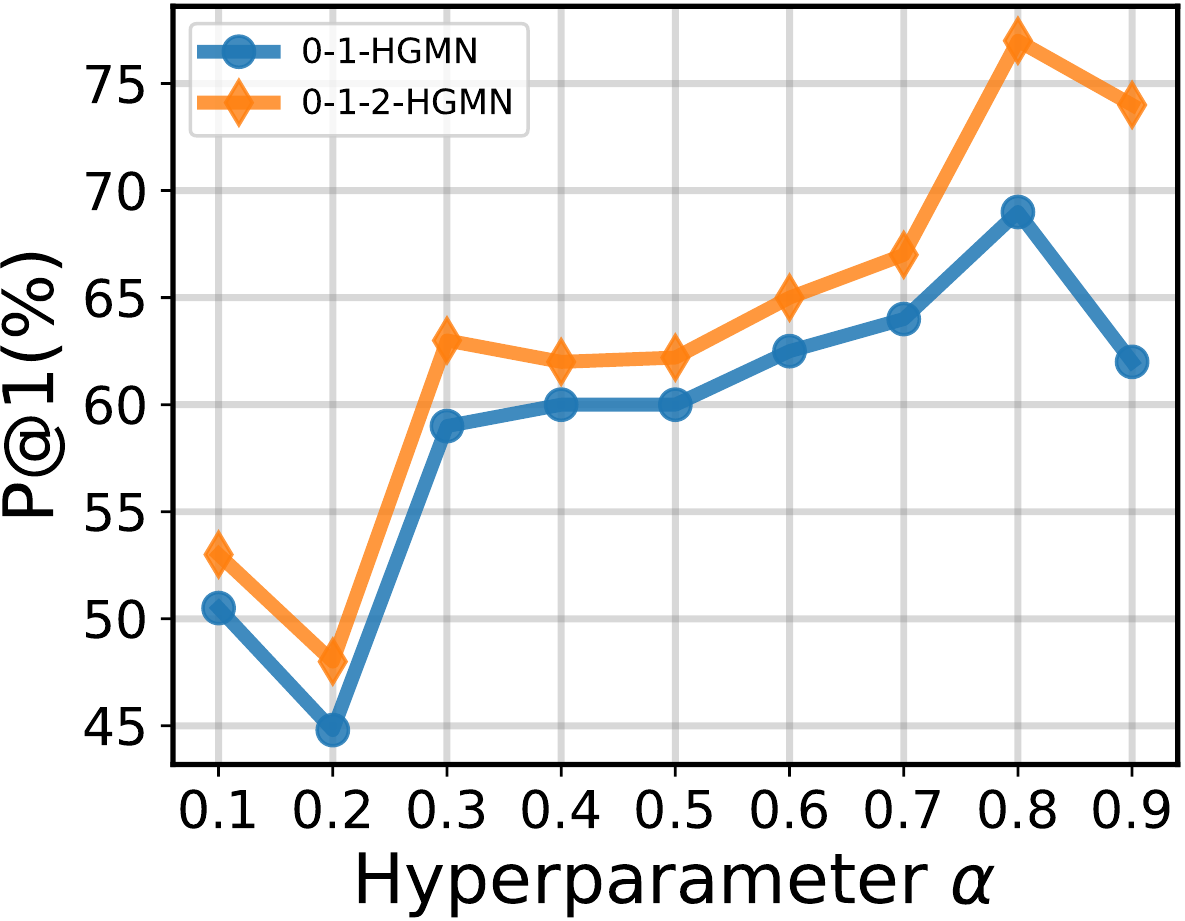}
}
\subfigure[Top$_{k}$ Sparsity]{
\centering
\includegraphics[width=0.23\linewidth]{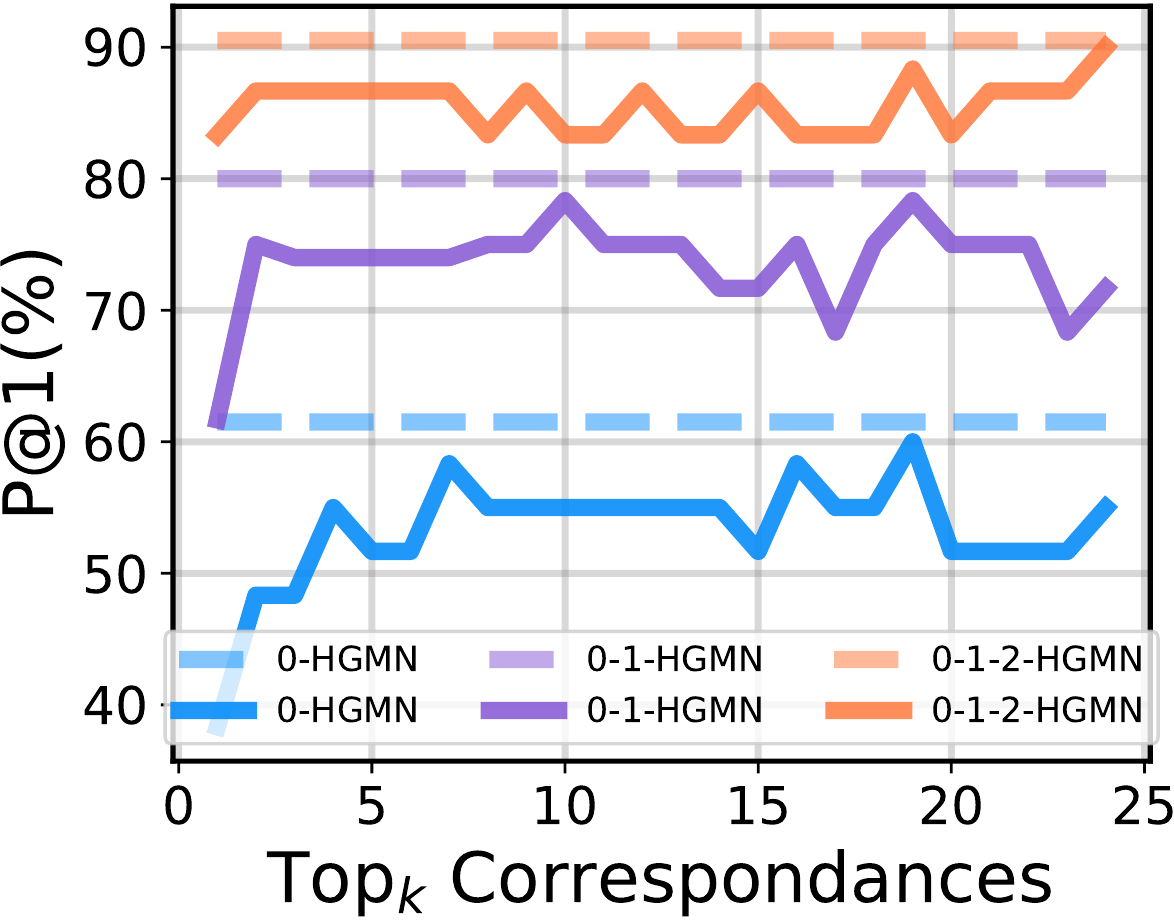}
}
\caption{Evaluation results of HGMN on synthetic datasets.}

\label{syn}
\end{figure*}

\begin{table*}[ht]
\renewcommand\arraystretch{1.2}
\centering
\caption{The $P@k$(\%) results for the training ratio Tr = $30\%$. Matching accuracies exceeding all baselines are marked in bold. The highest accuracies are underscored.}
\setlength{\tabcolsep}{4mm}{
\begin{tabular}{lccccccccc}
\hline

\hline
\multirow{2}{*}{\textbf{Method}}& \multicolumn{3}{c}{\textbf{Twitter-Foursquare}} & \multicolumn{3}{c}{\textbf{AI-DM}} & \multicolumn{3}{c}{\textbf{AI$_{13,14}$-AI$_{15,16}$}}\\
& \textbf{@1} & \textbf{@10} & \textbf{@30} & \textbf{@1} & \textbf{@10} & \textbf{@30} & \textbf{@1} & \textbf{@10} & \textbf{@30}\\
\hline
IONE\cite{liu2016aligning} & 3.3 & 18.6 & 31.4 & 5.7 & 26.4 & 36.6 & 7.5 & 30.1 & 43.2 \\
DeepLink\cite{zhou2018deeplink} & 2.1 & 11.5 & 23.0 & 1.8 & 16.6 & 28.9 & 2.7 & 23.6 & 35.1\\
CrossMNA\cite{chu2019cross} & 4.1 & 16.4 & 27.2 & 6.5 & 28.1 & 35.8 & 6.4 & 30.2 & 42.4\\
MGCN\cite{chen2020multi} & 1.2 & 7.0 & 12.9 & 1.5 & 15.5 & 21.7 & 1.9 & 15.6 & 26.1\\
DGMC\cite{fey2020deep} & 1.5 & 15.7 & 30.2 & 2.9 & 29.0 & 40.6 & 2.2 & 26.7 & 41.4 \\
\hline
CGRW & 1.2 & 11.8 & 23.3 & 2.5 & 26.9 & 41.6 & 0.4 & 22.5 & 39.2\\
GCN & 5.5 & 25.3 & 40.6 & 4.3 & 32.4 & 46.0 & 5.9 & 36.2 & 51.5\\
\hline\hline

1-HGMN & \textbf{8.2} & \textbf{29.6} & \textbf{44.4} & \textbf{6.4} & \textbf{32.5} & \textbf{46.8} & \textbf{9.3} & \textbf{37.2} & \textbf{51.9}\\
2-HGMN & \textbf{9.2} & \textbf{31.9} & \textbf{47.0} & \textbf{6.1} & \textbf{\underline{32.7}} & \textbf{47.1} & \textbf{9.9} & \textbf{37.2} & \textbf{51.8}\\

\hline

0-1-HGMN & \textbf{8.8} & \textbf{31.7} & \textbf{46.4} & \textbf{5.5} & \textbf{32.5} & \textbf{47.4} & \textbf{10.2} & \textbf{\underline{37.5}} & \textbf{\underline{52.1}}\\
0-1-2-HGMN & \textbf{\underline{10.1}} & \textbf{\underline{32.3}} & \textbf{\underline{48.3}} & \textbf{\underline{6.7}} & \textbf{\underline{32.7}} & \textbf{\underline{48.9}} & \textbf{\underline{10.4}} & \textbf{36.2} & \textbf{51.5}\\
\hline

\hline
\end{tabular}
}
\label{social}
\end{table*}

\begin{table*}[t]
\renewcommand\arraystretch{1.2}

\centering
\caption{The $P@k$(\%) results for DBP15K dataset. Matching accuracies exceeding all baselines are marked in bold. The highest accuracies are underscored.}
\scalebox{0.95}{
\begin{tabular}{lcccccccccccc}
\hline

\hline
\multirow{2}{*}{\textbf{Method}} & \multicolumn{2}{c}{\textbf{ZH$\to$EN}} & \multicolumn{2}{c}{\textbf{EN$\to$ZH}} & \multicolumn{2}{c}{\textbf{JA$\to$EN}} & \multicolumn{2}{c}{\textbf{EN$\to$JA}} & \multicolumn{2}{c}{\textbf{FR$\to$EN}} & \multicolumn{2}{c}{\textbf{EN$\to$FR}}\\
& \textbf{@1} & \textbf{@10} & \textbf{@1} & \textbf{@10} & \textbf{@1} & \textbf{@10} & \textbf{@1} & \textbf{@10} & \textbf{@1} & \textbf{@10} & \textbf{@1} & \textbf{@10}\\
\hline

\hline
GCN~\cite{wang2018cross} & 41.25 & 74.38 & 36.49 & 69.94 & 39.91 & 74.46 & 38.42 & 71.81 & 37.29 & 74.49 & 36.77 & 73.06 \\
B\footnotesize{OOT}\normalsize{EA}~\cite{sun2018bootstrapping}     & 62.94 & 84.75 & 60.98 & 81.29 & 62.26 & 85.39 & 58.25 & 83.10 & 65.30 & 87.44 & 61.79 & 85.08\\
M\footnotesize{U}\normalsize{GNN}~\cite{cao2019multi}      & 49.40 & 84.40 & 48.12 & 83.34 & 50.10 & 85.70 & 48.56 & 84.98 & 49.60 & 87.00 & 49.05 & 86.66 \\
NAEA~\cite{Zhu2019Neighborhood}       & 65.01 & 86.73 &       &       & 64.41 & 87.27 &       &       & 67.32 & 89.43 &       &      \\
RDGCN~\cite{wu2019relation}      & 70.75 & 84.55 & 67.73 & 84.53 & 76.74 & 89.54 & 74.53 & 89.24 & 88.64 & 95.72 & 86.80 & 95.41 \\
GMNN~\cite{xu2019cross}       & 67.93 & 78.48 & 65.28 & 79.64 & 73.97 & 87.15 & 71.29 & 84.63 & 89.38 & 95.25 & 88.18 & 94.75\\
DGMC~\cite{fey2020deep}       & 80.12 & 87.49 & 76.77 & 83.56 & 84.80 & 89.74 & 81.09 & 86.84 & \textbf{93.34} & 96.03 & 91.95 & 95.28\\
0-HGMN     & 78.89 & 91.84 & 73.92 & 88.67 & 81.01 & 93.44 & 80.74 & 93.12 & 91.85 & 97.74 & 90.61 & 97.51\\
\hline

\hline
1-HGMN     & \textbf{\underline{81.56}} & \textbf{93.46} & \textbf{\underline{78.09}} & \textbf{\underline{91.13}} & \textbf{\underline{85.05}} & \textbf{\underline{95.57}} & \textbf{\underline{82.75}} & \textbf{\underline{94.82}} & 93.20 & \textbf{\underline{98.46}} & \textbf{\underline{92.49}} & \textbf{\underline{98.13}}\\
2-HGMN     & \textbf{80.74} & \textbf{\underline{93.64}} & \textbf{76.93} & \textbf{90.85} & 84.17 & \textbf{95.46} & \textbf{82.08} & \textbf{94.19} & 92.10 & \textbf{98.43} & 91.52 & \textbf{98.07}\\

\hline

0-1-HGMN   & 79.07 & \textbf{92.34} & 75.30 & \textbf{89.70} & 81.77 & \textbf{94.29} & 79.15 & 92.84 & 91.62 & \textbf{97.96} & 90.15 & \textbf{97.77}\\
0-1-2-HGMN & 73.49 & 90.13 & 71.57 & 84.77 & 77.27 & 92.38 & 74.89 & 91.31 & 87.22 & 96.88 & 87.53 & 96.70\\
\hline

\hline
\end{tabular}
}
\label{dbp}
\end{table*}

\section{Experiments}
We verify our approach in three different settings. To avoid the cascaded expansion of high-order line graphs, we limit the highest order $m$ to $2$. $k$-HGMN\,($k\in\{0,1,2\}$) means HGMN is performed on the $k$-iterated line graph. In particular, 0-HGMN only utilizes information on original graph and thus can be considered as a variant of GNNs. In addition, 0-1-HGMN and 0-1-2-HGMN are the hierarchical variants with $m$=$1$ and $m$=$2$ respectively. We first demonstrate our method in an ablation study on synthetic graphs, and apply it to real-world tasks in social networks and cross-lingual knowledge graph alignment afterwards.

\subsection{Datasets and Metrics}
Our datasets include: \textbf{1) Twitter-Foursquare.} The two social networks are collected from Foursquare and Twitter \cite{zhang2015integrated}. \textbf{2) AI-DM.} Two co-author networks are extracted from papers published between 2014 and 2016 (collected by Acemap) in 8 representative conferences on Artificial Intelligence (AI) and Data Mining (DM) \footnote{AI conferences are IJCAI, AAAI, CVPR, ICCV, ICML, NeurIPS, ACL, and EMNLP, whereas the DM conferences include KDD, SIGMOD, SIGIR, ICDM, ICDE, VLDB, WWW, and CIKM.} respectively. \textbf{3) AI$\mathbf{_{13,14}}$-AI$\mathbf{_{15,16}}$.} Two co-author networks are constructed on papers published in 8 AI conferences (collected by Acemap) in 2013-2014 and 2015-2016 respectively. \textbf{4) DBP15K.} The datasets are generated from the multilingual versions of DBpedia \cite{sun2017cross}, which pair entities of the knowledge graphs in French, Japanese and Chinese into the English version and vice versa. All dataset statistics are listed in Table \ref{tab1}.

Following most works \cite{zhou2018deeplink,fey2020deep}, we adopt the standard metric $Precision$@$k$($P$@$k$) to evaluate the matching performance, which measures the proportion of correctly matched pairs ranked in the top $k$.

\subsection{Ablation Study on Synthetic Graphs}
We evaluate HGMN on synthetic graphs. We first construct an undirected Erdos-Renyi graph as the source graph $\mathcal{G}_s$ with $|\mathcal{V}_s|=100$ nodes and edge probability $p=0.1$, and a target graph $\mathcal{G}_t$ which is built from $\mathcal{G}_s$ by randomly deleting edges with probability $p_d$. For each $p_d\in \{0.0,0.1,0.2,0.3,0.4,0.5\}$, $100$ pairs of source and target graphs are generated and trained to report the average results.

\subsubsection{Architecture and parameters.}
We implement $\bm{\Psi}_{\theta_1}$ and $\bm{\Psi}_{\theta_2}$ by stacking $T$ layers of Graph Isomorphism Network (GIN) operator \cite{xu2018powerful}:
\begin{equation}
\small
\bm{x}_i^{(t)} = \mathrm{MLP}^{(t)}\Big((1+\epsilon^{(t)}) \cdot \bm{x}_i^{(t-1)}+\sum_{j \in \mathcal{N}(i)}\bm{x}_j^{(t-1)}\Big),
\end{equation}
considering its great power in distinguishing graph structures. Each MLP has a total number of layers $2$ and hidden dimensionality of $100$. The embedding for each node in $\mathcal{G}_s$ and $\mathcal{G}_t$ is initialized with one-hot encodings of node degrees. Following the setting of \cite{fey2020deep}, we apply ReLU activation and Batch normalization after each layer, and utilize \textit{Jumping Knowledge Style Concatenation} $\bm{x}_i=\bm{W}[\bm{x}_i^{(1)},\ldots,\bm{x}_i^{(T)}]$ to acquire the final node representation. We set $\alpha=0.9$ in Eq.~\ref{HGMN} and training ratio $T_r=0.7$ by default without specific mentioning.

\subsubsection{Results.}
Fig.~\ref{syn}(a) shows the matching accuracy P@1 for different structural noise $p_d$ with $T=3$. Benefiting from leveraging high-order and hierarchical structural information, 0-1-HGMN and 0-1-2-HGMN consistently outperform 0-HGMN. However, we observe that 1-HGMN and 2-HGMN are not always better than 0-HGMN. The difference is mostly caused by the initialization methods, as $k$-HGMN encodes the initial node features in $k$-ILG as combinations of the one-hot embeddings of corresponding nodes in the original graph, which is inferior than using the feature embeddings of ($k-1$)-HGMN. More specifically, 0-1-2-HGMN has better performance than 0-1-HGMN, but the improvement is much smaller than that from  0-HGMN to 0-1-HGMN. This reflects that the marginal benefit of learning high-order structural information is getting small with the increase of the order. Hence, a constant $k$ may exist that ($k-1$)-HGMN is superior to $k$-HGMN.

Fig.~\ref{syn}(b) visualizes the P@1 accuracy for different numbers of layers. It can be observed that the performance of 0-HGMN drops when $T > 9$, due to the overfitting and over-smoothing problem, which is in accords with the conclusion in  \cite{rong2019dropedge}. Conversely, the hierarchical variants acquire more stable P@1 scores as ILG incorporates high-order information such that deeper GIN obtains better performance.

Fig.~\ref{syn}(c) and \ref{syn}(d) show the P@1 accuracies for different values of hyperparameters. We set $T=3$ and $p_d=0.3$ in Fig.~\ref{syn}(c), $T=16$ and $p_d=0.3$ in Fig.~\ref{syn}(d). Fig~\ref{syn}(c) demonstrates that as the high-order information plays a growingly important part with the increase of $\alpha$, the accuracies increase except for $\alpha=0.9$. Fig.~\ref{syn}(d) shows the performance of HGMN on the $\rm{Top}_k$ correspondences. We observe that by choosing appropriate $k$s, the performance on the sparse correspondences is as good as that of the dense ones indicated by solid lines.

\subsection{Social Network Matching}
We evaluate HGMN on three real-world social networks: Twitter-Foursquare, AI-DM and AI$_{13,14}$-AI$_{15,16}$. All graphs are anonymous and only structural information can be used. Since HGMN is not embedding-based, we initialize each feature by \textit{Cross Graph Random Walk}\,(CGRW), which builds cross-network edges according to the ground truth and applys DeepWalk \cite{perozzi2014deepwalk} method.

\subsubsection{Architecture and parameters.}
GCN \cite{kipf2016semi} is adopted as our GNN operator:
\begin{equation}
\small
\bm{X}^{(t+1)}=\sigma\Big(\bm{\tilde{D}}^{-\frac{1}{2}}\bm{\tilde{A}}\bm{\tilde{D}}^{-\frac{1}{2}}\bm{X}^{(t)}\bm{W}^{(t)}\Big).
\end{equation}
Following each GCN layer, two MLP layers with $100$ hidden units are attached. We set $\alpha=0.5$, $T=3$ and the training ratio $Tr=0.3$. Moreover, the techniques of edge deletion and sparse correspondences are adopted. Hyperparameters include: $m = 2$, $d^{(0)}=10$, $d^{(1)}=5$, and $\rm{Top}_k=10$.

\subsubsection{Results.}
We report P@1, P@10 and P@30 accuracies of all methods in Table\,\ref{social}. HGMN achieves the highest P@k among all methods under different $k$s. In particular, HGMN outperforms those GCN-based methods, {\em i.e.,} DGMC and GCN\,(0-HGMN) with three-layer GCNs, which shows the effectiveness of our method and highlights the benefits of introducing high-order structural information. Although we use CGRW as the initialization method, it does not contribute to the high accuracy as the performance of CGRW is far inferior to HGMN. 

Across different variants of HGMN, 0-1-2-HGMN is more advantageous on Twitter-Foursquare and AI-DM dataset than on AI$_{13,14}$-AI$_{15,16}$. One possible reason is that the former two datasets have more complicated structures and heterogeneous attributes than the third one. When two corresponding nodes have more discrepant neighbors, more sophisticated methods are required to extract the local structures. Otherwise, the higher-order information would result in overfitting on graphs of homogeneous local structures. We also complement a series of experiments to identify the influence of Sinkhorn since some baselines are originally performed without Sinkhorn. The results are in the supplementary material.

\subsection{Cross-Lingual Knowledge Graph Alignment}
We also evaluate HGMN on the DBP15K dataset with directed and attributed graphs. We follow the setup of \cite{fey2020deep} by using the sum of word embeddings acquired by monolingual \textit{FASTTEXT} embeddings as the final entity input representation.

\subsubsection{Architecture and parameters.}
By referring to \cite{fey2020deep}, we implement GNN operator $\mathbf{\Psi}_{\theta_1}$ and $\mathbf{\Psi}_{\theta_2}$ as 
\begin{equation}
\small
\begin{aligned}
\bm{x}_i^{(t+1)} = \sigma \Big(\bm{W}_1^{(t+1)}\bm{x}_i^{(t)}+\sum_{j \in \mathcal{N}_{in}(i)}&\bm{W}_2^{(t+1)}\bm{x}_j^{(t)}\\ 
&+ \sum_{j \in \mathcal{N}_{out}(i)}\bm{W}_3^{(t+1)}\bm{x}_j^{(t)} \Big)
\end{aligned}
\end{equation}
where $\sigma$ is the ReLU activation function. 
We set $T=3$, $\alpha=0.5$, and $\rm{Top}_k=10$. And edge deletion are adopted with $d^{(0)}=5$ and $d^{(1)}=1$.

\subsubsection{Results.}
As shown in Table~\ref{dbp}, HGMN outperforms the state-of-the-art (including 0-HGMN) on all pairs of graphs with accuracy gain up to $7.95\%$. Our method is still superior to the GCN-based methods GCN and GMNN, while GCN and GMNN both adopt two-layer GCNs. Benefitting from reaching a data-driven neighborhood consensus between matched node pairs, DGMC surpasses GMNN and 0-HGMN in $P@1$. The hierarchical variants of HGMN are in general inferior to $k$-HGMN, and this may be because the source and target graphs of DBP15K share homogeneous local structures. The table also shows that $1$-HGMN has better performance than $2$-HGMN. This is consistent with the discussion in the ablation study where the higher-order ILG may not always bring better performance. Overall, the high-order structural information of directed graphs can be captured by ILGs and brings significant improvement.

\section{Conclusion}
To utilize high-order information on graphs, we propose a novel method called HGMN for graph matching. By introducing the iterated line graph in our framework, we leverage high-order information in a principled way and prove that our method is more expressive than GCNs in aligning corresponding nodes across graphs. Hierarchical variants of HGMN are also proposed to exploit the hierarchical representation of graphs. Evaluated on a variety of real-world datasets, HGMN has shown superior matching performance than the state-of-the-art.

\bibliographystyle{arxiv}
\bibliography{arxiv}

\end{document}